\documentclass[conference]{IEEEtran}
\IEEEoverridecommandlockouts
\usepackage{cite}
\usepackage{amsmath,amssymb,amsfonts}
\usepackage{algorithmic}
\usepackage{graphicx}
\usepackage{textcomp}
\usepackage{xcolor}
\def\BibTeX{{\rm B\kern-.05em{\sc i\kern-.025em b}\kern-.08em
    T\kern-.1667em\lower.7ex\hbox{E}\kern-.125emX}}
\begin{document}

\title{Prospects for Mitigating Spectral Variability in Tropical Species Classification Using Self-Supervised Learning\\
\thanks{The authors thank the Institut des Mathématiques pour la Planète Terre (IMPT) foundation for the first author’s PhD grant, Centre national d’´etudes spatiales (CNES) and Hytech-Imaging for the airborne imaging data and also the laboratory of excellence CEBA (Center for the Study of Bio-diversity in Amazonia,“Investissement d’Avenir” programs managed by Agence Nationale de la Recherche [CEBA, ref. ANR-10-LABX-25-01]) for the support of the research team. Authors declare no conflict of interest. Artificial intelligence systems were used to improve grammar and readability.}
}

\author{
\IEEEauthorblockN{Colin Prieur\IEEEauthorrefmark{1}\textsuperscript{*}, Nassim Ait Ali Braham\IEEEauthorrefmark{2}\IEEEauthorrefmark{3}, Paul Tresson\IEEEauthorrefmark{1}, Grégoire Vincent\IEEEauthorrefmark{1}, Jocelyn Chanussot\IEEEauthorrefmark{4}}
\IEEEauthorblockA{\IEEEauthorrefmark{1}AMAP, Univ Montpellier, CIRAD, CNRS, INRAE, IRD, Montpellier, France}
\IEEEauthorblockA{\IEEEauthorrefmark{2}Data Science in Earth Observation, Technical University of Munich, Munich, Germany}
\IEEEauthorblockA{\IEEEauthorrefmark{3}Remote Sensing Technology Institute, German Aerospace Center, Germany}
\IEEEauthorblockA{\IEEEauthorrefmark{4}Univ. Grenoble Alpes, Inria, CNRS, Grenoble INP, LJK, Grenoble, France}
\IEEEauthorblockA{*Corresponding author: colin.prieur@gmail.com}
}

\maketitle

\begin{abstract}
Airborne hyperspectral imaging is a promising method for identifying tropical species, but spectral variability between acquisitions hinders consistent results. This paper proposes using Self-Supervised Learning (SSL) to encode spectral features that are robust to abiotic variability and relevant for species identification. By employing the state-of-the-art Barlow-Twins approach on repeated spectral acquisitions, we demonstrate the ability to develop stable features. For the classification of 40 tropical species, experiments show that these features can outperform typical reflectance products in terms of robustness to spectral variability by 10 points of accuracy across dates.
\end{abstract}


\section{Introduction}
Remote sensing has become crucial for understanding the composition of tropical forests. It is essential for estimating biodiversity and planning sampling strategies that avoid expensive and potentially damaging field expeditions required for accurately identifying forest covers~\cite{laybros2020quantitative,ganivet2019towards}.
Thus, there is a need to transfer botanical expertise into a remote sensing context that can cover vast areas. In this regard, airborne hyperspectral imaging, as indicated in \cite{laybros2020quantitative,ganivet2019towards,colgan2012mapping}, emerges as a promising candidate for botanical identification. It provides precise radiometric information for species differentiation and provides extensive coverage, enabling large areas to be surveyed without ground-based efforts.

However, the precision and detail required for classification operations make these classification tools particularly sensitive to spectral variability~\cite{laybros2019across,prieur2024investigating}. 
For a fixed classification function, changes in reflectance due to sources of abiotic variability, such as atmospheric conditions, illumination, and variability of atmospheric correction, can severely affect the resulting spectrum~\cite{theiler2019spectral} and as such the performance of classifiers from one acquisition to another \cite{laybros2019across}.

To develop models capable of maintaining their classification performance across multiple acquisitions at different times or different places, two key questions must be answered:
\begin{itemize}
    \item Can spectral descriptors be encoded to remain resilient to variations in lighting, atmospheric conditions, and soil moisture during acquisition?
    \item Can classifiers maintain robustness when applied to different study sites that feature varying background reflectance, such as differing soil types, vegetation states (including phenological stages, hydric conditions), and varying degrees of canopy mixing, such as the abundance of lianas?
\end{itemize}

This article presents a deep learning approach for forest cover monitoring by developing reflectance-derived variables using a self-supervised learning (SSL) workflow, as implemented in previous work on standard HSI classification benchmarks~\cite{mou2017unsupervised,braham2024enhancing,liu2020deep,li2023few}. Although SSL methods have been applied in hyperspectral imaging to build relevant spectral representations~\cite{viel2023hyperspectral,mou2017unsupervised,braham2024enhancing}, their effectiveness under complex conditions in tropical forests has not yet been explored.

To address this challenge, we apply the Barlow-Twins method~\cite{zbontar2021barlow} with a unique construction of positive pairs of pixels from hyperspectral images captured on consecutive days. This approach aims to generate robust encodings against spectral variability due to atmospheric, illumination, and BRDF effects~\cite{vogtli2023effects,theiler2019spectral}.

We evaluated the robustness of these spectral variables using in situ canopy data from the Paracou in the French Guiana area and repeated hyperspectral overflights. A Linear Discriminant Analysis (LDA) model, applied to classify 40 tropical species, was trained on features from the first acquisition and tested on the second. Results show that SSL-generated features exhibit reduced variability compared to reflectance, as demonstrated by the fact that an LDA classification on SSL features achieves a 10-point higher robustness in classification accuracy across dates compared to the same operation on reflectance features.

The findings highlight the promise of this approach, particularly the effectiveness of minor spectral transformations, consistent with recent findings~\cite{moutakanni2024you}.

\section{Proposed Approach}

\subsection{Overview}\label{subsec:Overview}
Let $X_{T1}, X_{T2} \in \mathrm{R}^{N\times M \times C}$ be two hyperspectral cubes of identical resolution, mapping the same area, acquired at different times. We denote $x_{T_1}, x_{T_2} \in \mathrm{R}^{1\times 1 \times C}$ as two hyperspectral pixels of the same coordinate \textit{(i, j)} in matrices $X_{T_1}$ and $X_{T_2}$.

\begin{figure}[htbt]
    \centering
    \includegraphics[width=\linewidth]{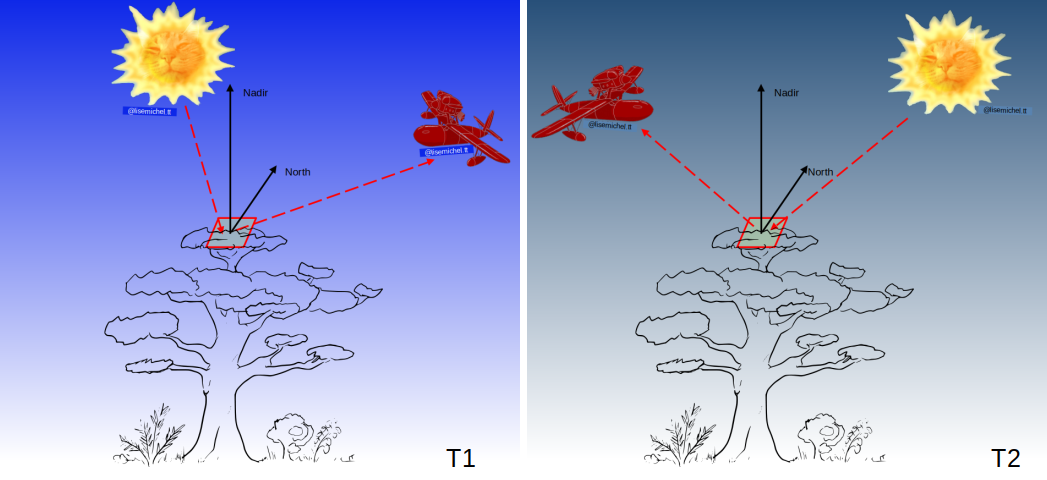}
\caption{Schematics of different physical process such as atmospheric variability, sun and view angle variability over a mapped coordinate of a canopy.}
\label{fig:DoubleView}
\end{figure}

As expressed in Fig.~\ref{fig:DoubleView}, across dates, many acquisition conditions can vary, leading to strong sources of variability over the spectra. The goal of this study is to train an encoder \textit{f} using these unlabeled pixels to build a more resilient spectral representation and to evaluate its robustness in classifying different tropical species over time.

\begin{figure}[htbt]
    \centering
    \includegraphics[width=\linewidth]{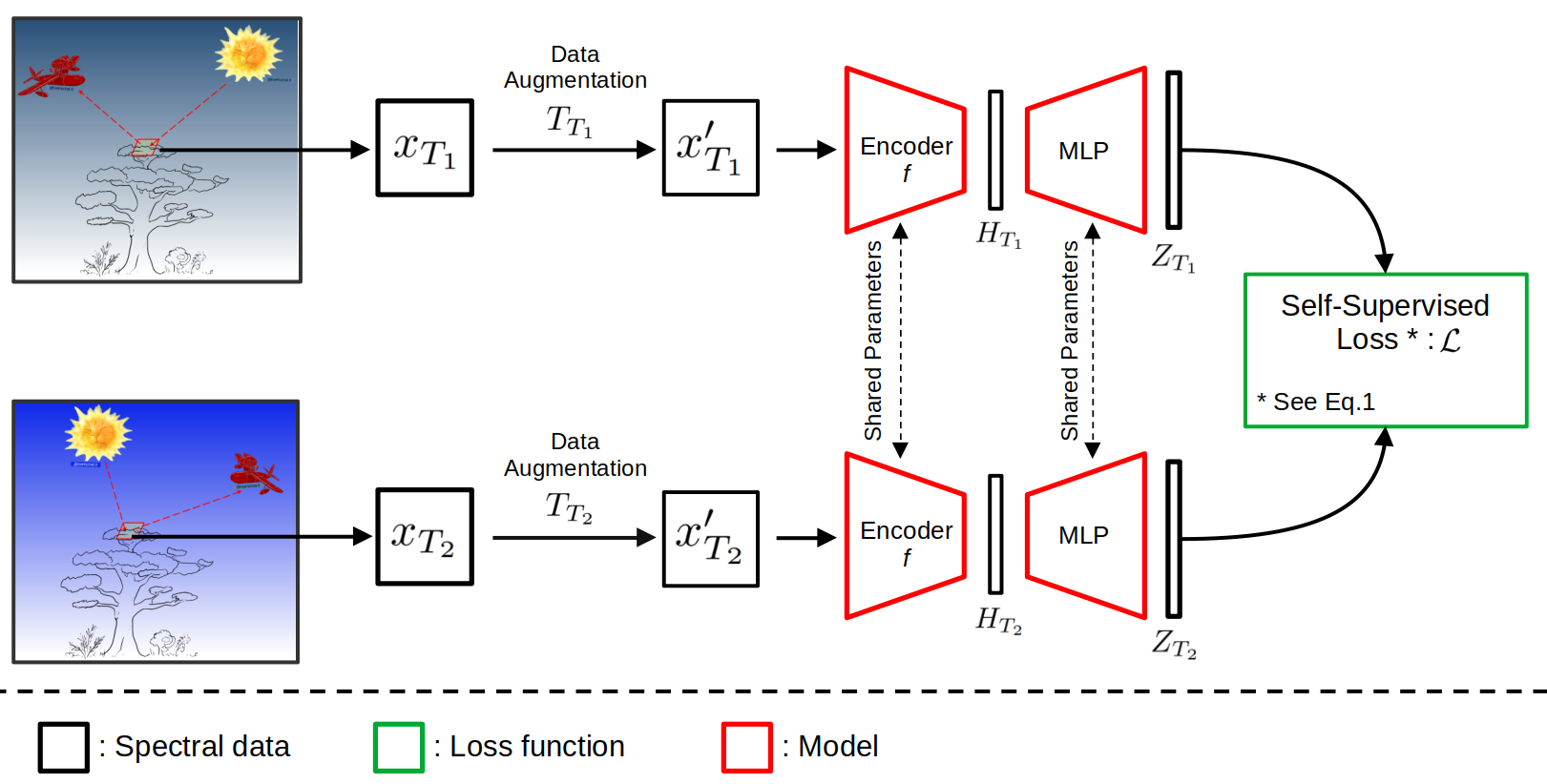}
\caption{Overall workflow for the Self-Supervised training across acquisitions.}
\label{fig:Overview}
\end{figure}

The method involves two steps: pre-training the encoder \textit{f} as illustrated in Fig.~\ref{fig:Overview}, followed by training a Linear Discriminant Analysis (LDA) model using the encoder's representations at one date and testing it on the other date on labeled spectra extracted from 40 different tropical species outlines of crowns; see section.~\ref{subsubsection:GroundTruth}. This approach, common in SSL for hyperspectral data~\cite{braham2024enhancing,mou2017unsupervised}, uniquely incorporates temporal variability on the spectrum to improve the robustness of classification.

\subsection{Pretraining Stage}
For pre-training, as described in~\cite{braham2024enhancing}, we treat all pixels as unlabeled and train an encoder \( f \) using the Barlow-Twins algorithm, described below. Barlow-Twins is an SSL method based on redundancy minimization. From a coordinate \textit{(i, j)}, two views \( x_{T_1}' = T_{T_1} (x_{T_1}) \) and \( x_{T_2}' = T_{T_2} (x_{T_2}) \) are stochastically generated using two sets of augmentations \( T_{T_1} \) and \( T_{T_2} \). In this work, we leverage temporal information to create the views by sampling pixels from the same coordinates at different times (see Section~\ref{subsec:Overview}). These views are fed into a shared encoder \( f \) (see section~\ref{subsection:Models&Metrics}) to obtain their representations \( H_{T_1} = f (x_{T_1}') \) and \( H_{T_2} = f (x_{T_2}') \). An additional Multi-Layer Perceptron (MLP) head is then applied to obtain \( Z_{T_1} = MLP (H_{T_1}) \) and \( Z_{T_2} = MLP (H_{T_2}) \). Since \( x_{T_1}' \) and \( x_{T_2}' \) are augmented views and thus semantically identical, their representations should be similar.

Moreover, according to the principle of redundancy minimization, the dimensions of \( Z_{T_1} \) and \( Z_{T_2} \) should be decorrelated. These properties can be enforced on a batch level by optimizing the following loss:

\begin{equation}
    \mathcal{L} = \sum_{k}(1 - C_{kk})^{2} + \lambda \sum_{k}\sum_{l \neq k} C_{kl}^{2}
\end{equation}

Where \( \lambda \in \mathbb{R}^+ \) is a trade-off parameter and \( C \) is the cross-correlation matrix computed between the outputs \( Z_{T_1} \) and \( Z_{T_2} \) of the network branches along the batch dimension:

\begin{equation}
    C_{kl} = \frac{\sum_b z_{b,k}^{T_1} z_{b,l}^{T_2}}{\sqrt{\sum_b (z_{b,k}^{T_1})^{2}} \sqrt{\sum_b (z_{b,l}^{T_2})^{2}}}
\end{equation}

\subsection{Classification Stage}

Based on the encodings generated during the pre-training stage, we aim to determine the relevance and robustness of these representations to changes in acquisition conditions.

To achieve this, we use encodings produced by the pre-trained model with fixed weights. Using these encodings, we train a Linear Discriminant Analysis (LDA) on labeled spectra of tropical trees extracted from $X_{T_1}$ and test its performance on spectra with the same labels, extracted from $X_{T_2}$.

It is important to note that the classification operation serves merely as a downstream task designed to evaluate the stability of the feature space, rather than to demonstrate the relevance of self-supervised learning to improve model classification performance~\cite{braham2024enhancing}. 

Our focus is on comparing the stability of classification performance over time with that of raw reflectance, across various view pairing strategies.

\subsection{Views generation}
\subsubsection{Pair Sampling}

Pixels with the same coordinate between acquisitions, which are closely time-related, capture a temporal variability in spectral reflectance of the same object (tree crown) but resulting from different acquisition conditions, as illustrated in Fig.~\ref{fig:DoubleView}.

This sampling strategy allows the integration of regularization based on the sum of abiotic physicochemical processes affecting spectral variability; see Fig.~\ref{fig:DoubleView}. These include atmospheric composition variability, Bidirectional-Reflectance function effects, and the inherent variability arising from uncertainties in atmospheric correction~\cite{vogtli2023effects,theiler2019spectral}.

As stated in~\cite{braham2024enhancing}, this particular sampling is applied to reduce the need for aggressive or naive transformation as applied in SSL, often different from a hyperspectral airborne imaging context. 
Ultimately, it is possible to construct 2.3 million different views composed of pairs of pixels from the same coordinates extracted from two different acquisition dates for the Paracou area, see section.~\ref{Subsection:Area_Acquisition}.

\subsubsection{Data Augmentation}
Only a few transforms were considered, restricted to spectral augmentations. The scaling transform was used to mimic the simple effects of spectral variability~\cite{drumetz2019spectral,hapke2012theory,theiler2019spectral}, where the others were used given their positive impact on other SSL workflows on hyperspectral data~\cite{braham2024enhancing}. 

Used transforms are Random Band Swapping of adjacent bands, Additive Gaussian Noise, and a randomly defined multiplicative factor applied to the SWIR and VNIR domain of each augmented spectra, as inspired by \cite{drumetz2019spectral} from \cite{hapke2012theory} called Scaling Domain. 

\section{Experimental Setting}
\subsection{Data}
\subsubsection{Study Area and Hyperspectral Acquisition}\label{Subsection:Area_Acquisition}
The Paracou research station, located in northern French Guiana (5°16’ N, 52°55’ W), approximately 15 km from the coast, experiences an annual average temperature of 26°C with a variation of ±1°C and annual rainfall of around 2875 ± 510 mm. The site's altitude ranges from 5 m to 45 m above sea level.
Hyperspectral data was acquired by Hytech-Imaging using two sensors: Hyspex VNIR-1600 and SWIR-384me. The VNIR sensor covers 414 to 993 nm in 160 spectral bands with 3.7 nm spectral sampling, while the SWIR sensor covers 976 to 2512 nm in 288 spectral bands with 5.45 nm spectral sampling. The Ground Sampling Distance (GSD) was 1 m for the VNIR and 2 m for SWIR data.
Overflights of the Paracou site occurred on September 19 and 20, 2016. The first acquisition ran from 1:12 PM to 3:38 PM local time, starting on the western side of the site. The second acquisition, which covered a smaller area, took place between 2:14 pm and 3:00 pm local time.

Hyperspectral acquisitions were then corrected using the Parge/Atcor-4 for flat terrain software~\cite{Richter2023atcor,schlapfer2002parge} and then co-registered with the Arosics software~\cite{AROSICS2017} to improve co-registration accuracy. Finally, post-processing corrections were applied to prune noisy bands and interpolate overly corrected bands, resulting in a concatenated array covering four spectral domains, which are : 
\begin{itemize}
    \item The VNIR domain, [414.7nm-975.0nm], 154 bands
    \item The SWIR0 domain, [976.9nm-1329.7nm], 66 bands
    \item The SWIR1 domain, [1497.9nm-1774.8nm], 52 bands
    \item The SWIR2 domain, [1981.0nm-2361.0nm], 71 bands
\end{itemize}

\subsubsection{Ground-Truth}\label{subsubsection:GroundTruth}

A continuous field survey is conducted in Paracou to create a large ground-truth data set. Within this field-checked dataset, 40 species, totaling 1,376 trees, were selected. Only crowns mapped on both dates were considered. The abundances of tropical species are typically highly imbalanced~\cite{ter2013hyperdominance}; in our dataset, the least represented species has only three trees, while the most represented species has 245 trees. This imbalance affects the total number of reflectance spectra per species, but we chose to retain the natural representation without balancing any class.

\subsection{Models and Metrics}\label{subsection:Models&Metrics}
This study employs a simple 1D convolutional neural network model comprising five convolutional layers without residual blocks, based on~\cite{viel2023hyperspectral}. The MLP used at the output of this network consists of two fully connected layers, designed according to~\cite{zbontar2021barlow}, with an output dimension of 2056 parameters.

For each experiment, the Barlow-Twins model was trained for 30 epochs using the Adam optimizer.

The performance of encoding was measured using the average classification accuracy between species as the metric in the test data set derived from acquisitions at time T2. The registered mean accuracy is the best output on the test data set between checkpoints.

\section{Experimental Results}
\begin{figure}[htbt!]
    \centering
    \includegraphics[width=1\linewidth]{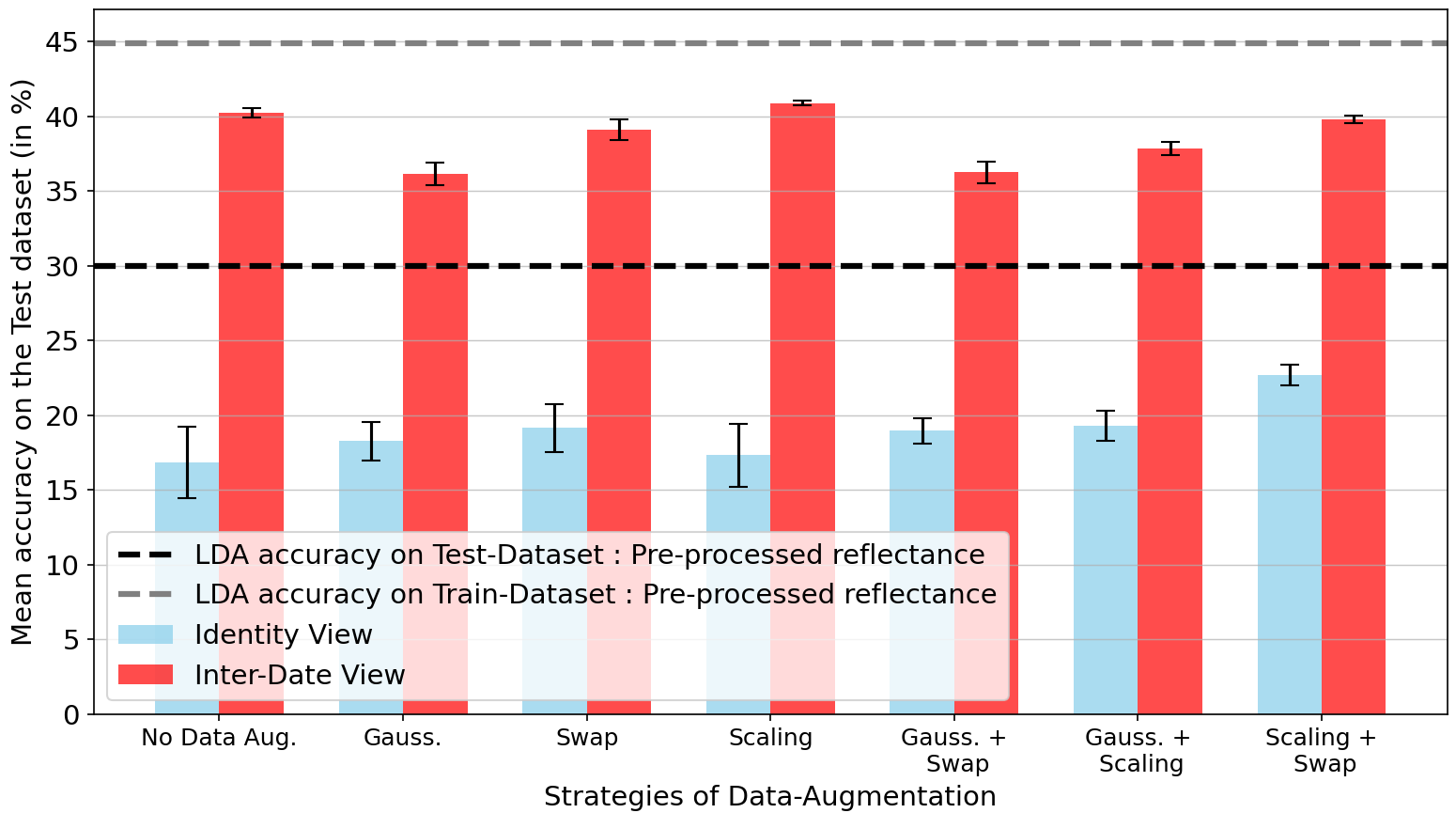}
\caption{Test dataset mean classification accuracy for 40 tropical species for each SSL pre-training strategy compared to LDA results on pre-processed reflectance. For each data-augmentation strategy (x-axis), red bars show accuracy with the positive pair construction strategy, and blue bars without pairing. Bars indicate mean accuracy across all runs, with uncertainty intervals showing standard deviation from four independent runs. Dotted grey and black lines mark the average classification accuracy  on the training and test datasets, respectively.}
\label{fig:Performance}
\end{figure}

By aggregating the accuracy values recorded by the LDA on the test dataset across multiple experiments for the same positive pair construction strategy, we report the LDA's performance variations with different view pairing and data augmentation techniques; see Fig.~\ref{fig:Performance}. 

The uncertainty intervals presented in Fig.~\ref{fig:Performance} represent the standard deviation computed on 4 different runs by self-supervised strategies.

\subsection{Impact of Views-pairing}
As shown in Fig.~\ref{fig:Performance}, only inter-date view pairing appears to significantly enhance the consistency of features across date by at least 5 points of accuracy compared to the baseline reflectance across all data-augmentation strategies. This outcome is expected since the perturbations between the training and test datasets are the same as those presented in the views. This strategy, which could be interpreted as a Self-Supervised calibration, significantly minimizes the performance loss due to changes in acquisition dates.

Without pair construction, performance of trained features is invariably suboptimal and below baseline reflectance LDA classification accuracy by at least 7 points of accuracy across all data-augmentation strategies. Thus, in that case, the reflectance is more robust than the output of a Barlow-Twins model trained on identical pairs. This outcome is also expected, as without pair construction, the provided pairs are similar or identical to each other and do not introduce any relevant perturbation that could properly represent a variability due to a change of date. However, data-augmentation were shown to induce slightly more robust resilience to trained features, their influence is described in section~\ref{sec:DataAugmentation}.

In all settings, the results in Fig.~\ref{fig:Performance} indicate that pairing through co-acquisitions remains a promising solution to reduce the sensitivity of acquisitions to changing atmospheric conditions.

\subsection{Impact of Data Augmentation}\label{sec:DataAugmentation}
Fig.~\ref{fig:Performance} shows that, although the impact of data augmentation is not as strong as positive pair construction, it still affects the performance of features resilience. Without pairing strategies, data augmentations are the only source of additional variability, helping reduce classification performance degradation across dates by a moderate amount, or are even detrimental in the case of the single scaling transform strategy. They do not yield more stable representations than pre-processed reflectance. 

More aggressive transformations could potentially produce more robust encodings; however, this introduces a trade-off between the resilience of features and their relevance to the downstream task, reducing their effectiveness in the complex task of classifying 40 tropical tree species. Therefore, effective transformations remain mild and cannot match the variability to be addressed between dates.

Across all pairing strategies, the best test performance is achieved through domain scaling transformations combined with other perturbations, be it pairing strategies or other data-augmentation techniques, which align with common spectral variability as discussed in~\cite{drumetz2019spectral,hapke2012theory,theiler2019spectral}. In terms of sensitivity, scaling consistently yields strong results regardless of the severity of the transformation. On the other hand, the effectiveness of Gaussian noise heavily depends on the noise variance, with optimal performance at 0.5\% of reflectance, though it still remains less effective than other transformations.

\section{Conclusion}
We show that SSL makes it possible to encode spectral descriptors that are resilient to changes in abiotic conditions during acquisition by building positive pairs of pixels over the same coordinate across overflights. Although further work is needed to stabilize classifier performance across different areas or acquisition dates, an SSL approach using carefully constructed positive pairs shows promise. By emphasizing extrinsic spectral variability (unrelated to canopy properties) under the strong assumption of perfect acquisition alignment, this approach can help build more resilient and relevant spectral descriptors.

\bibliographystyle{ieeetr}
\bibliography{references.bib}

\end{document}